\documentclass[letterpaper,10pt,conference]{IEEEconf}  

\IEEEoverridecommandlockouts                              

\usepackage{graphics} 
\usepackage{epsfig} 
\usepackage{amsmath} 
\usepackage{amssymb}  
\usepackage{url}
\usepackage{multirow}

\usepackage{glossaries}
\setacronymstyle{long-short}

\title{\LARGE \bf
Robotic Arm Platform for Multi-View Image Acquisition and 3D Reconstruction in Minimally Invasive Surgery}

\author{Alexander Saikia$^{1}$, Chiara Di Vece$^{1}$, Sierra Bonilla$^{1}$, Chloe He$^{1}$, Morenike Magbagbeola$^{1}$,\\ Laurent Mennillo$^{1}$, Tobias Czempiel$^{1,2}$, Sophia Bano$^{1}$ and Danail Stoyanov$^{1}$
\thanks{*This work was supported in whole, or in part, by the Wellcome/EPSRC Centre for Interventional and Surgical Sciences (WEISS) [203145/Z/16/Z], the Department of Science, Innovation and Technology (DSIT) and the Royal Academy of Engineering under the Chair in Emerging Technologies programme; EPSRC Centre for Doctoral Training in Intelligent, Integrated Imaging In Healthcare (i4health) [EP/S021930/1]. For the purpose of open access, the author has applied a CC BY public copyright licence to any author accepted manuscript version arising from this submission.}
\thanks{$^{1}$A. Saikia, C. Di Vece, S. Bonilla, C. He, L. Mennillo, T. Czempiel, S. Bano and D. Stoyanov are  with the UCL Hawkes Insitute, the Department of Medical Physics and Biomedical Engineering and the Department of Computer Science, University College London, London, United Kingdom}%
\thanks{$^{2}$ T. Czempiel is with EnAcuity Ltd., London, United Kingdom.}
\thanks{Corresponding author: {\tt\small alexander.saikia.21@ucl.ac.uk}}
}

\begin{document}

\newacronym{mis}{MIS}{minimally invasive surgery}
\newacronym{ramis}{RAMIS}{robot-assisted minimally invasive surgery}
\newacronym{lg}{LG}{LightGlue}
\newacronym{2d}{2D}{2-dimensional}
\newacronym{3d}{3D}{3-dimensional}
\newacronym{sfm}{SfM}{Structure from Motion}
\newacronym{slam}{SLAM}{Simulataneous Localisation and Mapping}
\newacronym{sift}{SIFT}{Scale Invariant Feature Transform}
\newacronym{hd}{HD}{high-definition}
\newacronym{rcm}{RCM}{remote centre of motion}
\newacronym{ik}{IK}{inverse kinematic}
\newacronym{dof}{DoF}{Degrees of Freedom}
\newacronym{icp}{ICP}{iterative closest point}
\newacronym{rmse}{RMSE}{Root Mean Squared Error}
\newacronym{rgb}{RGB}{Red-Green-Blue}
\newacronym{ct}{CT}{computerised tomography}
\newacronym{rpe}{RPE}{relative pose error}
\newacronym{ompl}{OMPL}{Open Motion Planning Library}
\newacronym{sdk}{SDK}{Software Development Kit}
\newacronym{fov}{FoV}{field of view}

\maketitle
\thispagestyle{empty}
\pagestyle{empty}

\begin{abstract}

Minimally invasive surgery (MIS) offers significant benefits such as reduced recovery time and minimised patient trauma, but poses challenges in visibility and access, making accurate 3D reconstruction a significant tool in surgical planning and navigation. This work introduces a robotic arm platform for efficient multi-view image acquisition and precise 3D reconstruction in MIS settings.
We adapted a laparoscope to a robotic arm and captured ex-vivo images of several ovine organs across varying lighting conditions (operating room and laparoscopic) and trajectories (spherical and laparoscopic). We employed recently released learning-based feature matchers combined with COLMAP to produce our reconstructions. The reconstructions were evaluated against high-precision laser scans for quantitative evaluation.
Our results show that whilst reconstructions suffer most under realistic MIS lighting and trajectory, many versions of our pipeline achieve close to sub-millimetre accuracy with an average of 1.05~mm Root Mean Squared Error and 0.82~mm Chamfer distance. Our best reconstruction results occur with operating room lighting and spherical trajectories.
Our robotic platform provides a tool for controlled, repeatable multi-view data acquisition for 3D generation in MIS environments which we hope leads to new datasets for training learning-based models.
\end{abstract}


\section{INTRODUCTION}
Surgery has experienced remarkable evolution, particularly with the rise of minimally invasive techniques like endoscopy and laparoscopy. These methods have transformed modern surgical practice by reducing patient recovery time and minimizing tissue damage. However, they also present challenges, such as limited visibility, reduced tactile feedback, and increased cognitive demands on surgeons. As surgical techniques continue to advance, there is an increasing need for innovative technologies to address these challenges, enhance precision, and improve patient outcomes~\cite{Maier-Hein2017}.\\
In \gls{mis} and especially \gls{ramis}, endoscopy is crucial for navigating the surgical field.
Moreover, accurate and reliable \gls{3d} reconstruction of organs during surgery is vital for advancing downstream computer-assisted tasks, such as augmented and virtual reality~\cite{NICOLAU2011189, Chong2022} and enhanced visualisation for surgical navigation~\cite{overley2017}.
Tools accomplishing these downstream tasks can augment the surgeon's understanding of the operating field and enable the identification of key structures such as blood vessels and tumours, improving both safety and precision in \gls{mis} and \gls{ramis}~\cite{Bianchi2020}.
However, the process of producing \gls{3d} reconstructions is challenging owing to the complexity of the surgical environment and the limited availability of image data with adequate \gls{3d} ground truth.


\begin{figure}
    \centering
    \includegraphics[width=\linewidth]{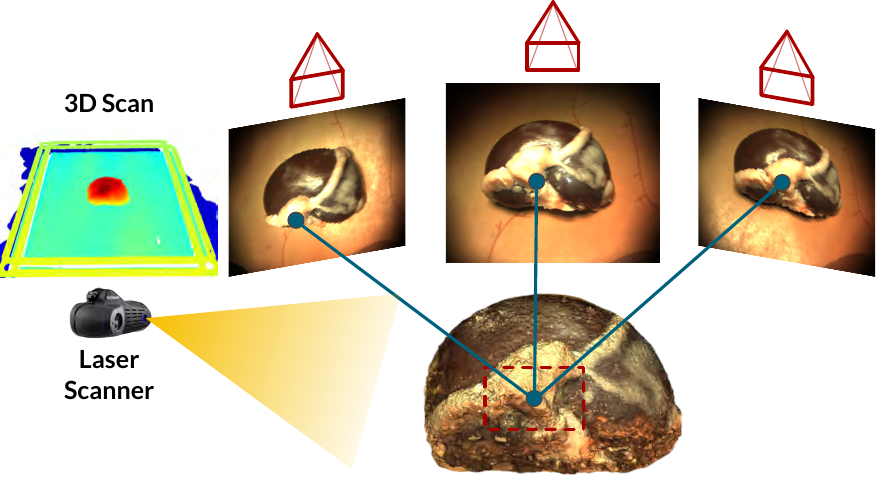}
    \caption{This work creates \gls{3d} reconstructions from multi-view images collected by our robotic arm platform. We use a laser scanner to obtain ground truth data for comparison.}
    \label{fig:are-you-not-motivated}
\end{figure}

\begin{table*}
\centering
\caption{Table Summary of publicly available datasets in surgical domain with ground truth properties and size.}
\label{tab:relwork}
\begin{tabular}{lllll}
\hline
\textbf{Dataset} & \textbf{System \& Imaging Modality} & \textbf{Ground Truth Modality} & \textbf{Scenes} & \textbf{Size (frames)} \\ 
\hline
\textbf{CV3D}~\cite{bobrow2023} & Synthetic Colonoscopy & Synthetic & Simulated colonoscopy scenes & 30073 \\ 
\hline
\textbf{SCARED}~\cite{allan2021stereo} & Stereo (da Vinci Xi) & Structured Light Depth & Porcine Abdomen & 17607 \\ 
\hline
\textbf{SERV-CT}~\cite{edwards2022serv} & Stereo (da Vinci) & CT-based Segmentation & Porcine liver, kidney, heart & 32 \\ 
\hline
\textbf{StereoMIS}~\cite{hayoz2023learning} & Stereo (da Vinci Xi) & Stereo Depth Estimation & Porcine and human & 14804 \\ 
\hline
\textbf{Ours} & Multi-view (Kinova Gen 3) & Laser Scans & Ovine liver, kidney & 29806 \\ 
\hline
\end{tabular}%
\end{table*}

This work presents a robotic arm-based platform that addresses these challenges by enabling multi-view image acquisition and \gls{3d} reconstruction in MIS environments. By utilising a robotic arm we ensure more consistent, reliable scanning as well as obtaining pose information. The system is designed to accommodate various imaging modalities and employs a laser scanner to capture highly accurate \gls{3d} ground truth, allowing for comprehensive evaluation of the reconstruction pipeline as seen in Figure~\ref{fig:are-you-not-motivated}. Our approach utilises state-of-the-art feature matching methods, such as ALIKED~\cite{zhao2023aliked} and GIM~\cite{shen2024gim}, paired with \gls{lg}~\cite{lindenberger2023lightglue} for robust correspondence across frames, and COLMAP~\cite{schoenberger2016sfm,schoenberger2016mvs} for dense \gls{3d} reconstruction. The platform is validated by the collection and processing of multiple \textit{ex-vivo} organ datasets, and its performance is benchmarked against ground truth laser scans to demonstrate the accuracy and robustness of the proposed system.
\newpage
Our main contributions and \gls{3d} reconstruction pipeline for surgery can be summarised as:
\begin{itemize}
    \item We customise a robotic arm platform adding a laparoscopic system suitable for medical applications;
    \item We demonstrate our acquisition protocol with different design choices such as trajectories and lighting to account for high realism and adaptability to the medical domain;
    \item We show how the multi-view images taken by the system can be used to extract \gls{3d} reconstructions using established algorithm choices and comparing them with each other both qualitatively and quantitatively using relevant metrics.
\end{itemize}
Our platform represents a step towards semi-automatic \gls{3d} reconstruction of organs, which has the potential of being translated to surgical vision enhancement such as registration of pre-operative data such as MRI or PET, surgical measurements and advanced visualisation and guidance to aid the surgeon whereby improving patient outcome. 

\section{RELATED WORK}

\subsection{Datasets}
The field of \gls{ramis} has seen significant advancements, improving the safety, ease, and effectiveness of procedures~\cite{howe1999robotics, vitiello2012emerging, attanasio2021autonomy}.
The use of \gls{3d} visualisation in \gls{mis} to enhance surgical views has been shown to be beneficial for outcomes~\cite{wagner2012three}. Therefore to reap this benefit, high quality data is essential for the development of algorithms for \gls{ramis}~\cite{haidegger2022robot}. However, despite the growing use of \gls{ramis}, publicly available datasets with robot kinematics/poses and ground truth \gls{3d} information are rare due to logistical and ethical constraints. 
Current datasets often suffer from significant limitations in terms of size, realism and ground-truth accuracy, leading to concerns about their ability to generalise. Table~\ref{tab:relwork} includes a list of commonly used datasets.

For example, CV3D~\cite{bobrow2023} is a synthetic colonoscopy dataset comprising 22 short video sequences generated using phantom data. 
Given the synthetic nature of the dataset, the additional \gls{3d} data is accurate and reliable; however, the dataset lacks anatomical realism which poses challenges when training deep learning models.

The most commonly used dataset is SCARED~\cite{allan2021stereo}, a stereo dataset obtained using a da Vinci Xi surgical robot. 
The ground truth was produced by obtaining 4 or 5 keyframes using structured light. 
Using the kinematics available from the robot, approximate depth maps were obtained by reprojecting depth maps from a keypoint.
The SCARED dataset has time synchronization issues between the \gls{rgb} video and kinematics, as well as ground truth depth misalignment with the \gls{rgb} data.
Some data has inaccurate intrinsic calibrations, which will affect point cloud reprojections and stereo rectification.

The SERV-CT~\cite{edwards2022serv} dataset is a validation dataset that offers 16 stereo-endoscopic image pairs acquired using the da Vinci surgical robot, each with \gls{ct}-based anatomical segmentations and occlusion maps. 
The comparatively small sample size, limited to 16 pairs, increases the risk of overfitting if used as the primary training dataset.

StereoMIS~\cite{hayoz2023learning} is an in-vivo dataset recorded using a da Vinci Xi surgical robot consisting of 3 porcine and 3 human subjects. Similar to SCARED, the ground truth poses are derived from the da Vinci's forward kinematics. While no ground truth depth data is available, the stereo images can be processed, and the single camera stream can be used to investigate monocular techniques. 
In creating a new platform for dataset acquisition we aim to address the significant complexity in creating a new surgical \gls{3d} dataset.

\subsection{3D Reconstruction for Surgery}
After the collection of multi frame surgical data, \gls{3d} reconstruction is essential to enhance the surgeon's view and proceed with downstream tasks. There are numerous \gls{3d} reconstruction algorithms, including \gls{sfm}, \gls{slam} and, if a stereo endoscope is used, stereo reconstruction~\cite{maier2014comparative,xu2024review}. These algorithms rely on feature extraction and matching between \gls{2d} images for generating \gls{3d} point clouds, which are used to estimate depth and camera pose. Traditional feature extraction and matching methods like SIFT~\cite{lowe2004sift} have been widely used for this purpose. However, in surgical environments, particularly when dealing with low-texture organs, more sophisticated methods can generate more accurate matching.

In~\cite{bonilla2024mismatched}, the authors demonstrated that combinations of ALIKED with \gls{lg} and GIM with \gls{lg} yielded the best results on out-of-domain data. Other learning-based detectorless techniques, such as Dust3r~\cite{dust3r_cvpr24} and Mast3r~\cite{mast3r_arxiv24} process the entire pipeline at once, making them highly computationally intensive and impractical for large datasets on standard machines.

\section{Robotic Platform for 3D Generation}
\begin{figure*}
     \centering
     \includegraphics[width=1.0\textwidth]{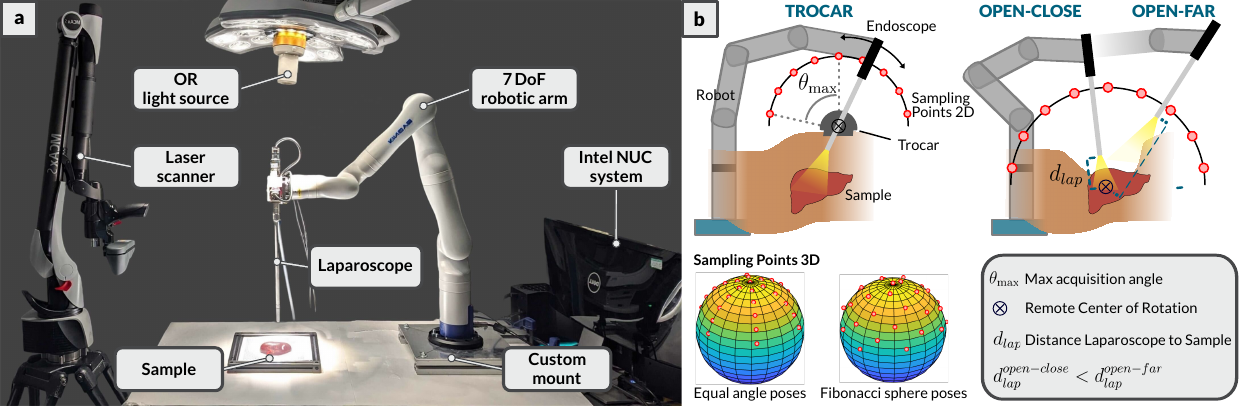}
     \caption{\textbf{a)} Annotated figure showing the robotic platform. The robot is attached to the table using a custom mount. The laparoscope optical system is attached to the robot with \gls{3d} printed and laser-cut parts. All trajectory control and data capture are handled by an Intel NUC. \textbf{b)} Depiction of acquisition setup and the three trajectories: Trocar, Open-Close and Open-Far. Additionally, the sampling points in \gls{3d} are depicted for both equal angle poses where poses are calculated by sequentially incrementing the azimuth and altitude angles by fixed amounts and Fibonacci sphere poses which are more even coverage over the imaging sphere.}
     \label{fig:platform}
 \end{figure*}

This section outlines our robotic arm platform for multi-view image acquisition and \gls{3d} reconstruction in \gls{mis}, as seen in Figure~\ref{fig:platform}. We describe the design of the robotic arm, imaging and recording systems, introduce the various scanning trajectories implemented using the robot and discuss our hand-eye calibration method.
\subsection{Robotic Arm}
 The platform utilises a Kinova Gen 3 7-\gls{dof} robotic arm\footnote{\url{www.kinovarobotics.com}}. The arm has a maximum reach of 902~$mm$ and a maximum speed of 0.5~$m/s$ at the end effector. Unlike many other robotic arms, the actuators can rotate infinitely, allowing more of the workspace to be covered by the Kinova making it ideal for our application of capturing the different organs from different angles. \\
 We interface with the robot using a modified version of ros2-kortex package\footnote{\url{https://github.com/Kinovarobotics/ros2_kortex}}. The platform uses the robotic manipulation platform MoveIt2\footnote{\url{https://moveit.ai}}~\cite{coleman2014reducing} for motion planning and kinematics control. A custom \texttt{moveit\_config} was created to suit our robot-laparoscope configuration.

\subsection{Imaging}
Our imaging system uses a FLIR Blackfly S USB3 (BFS-U3-50S5C-C, Teledyne FLIR, Wilsonville, Oregon, USA) \gls{rgb} camera custom mounted to a (HOPKINS\textregistered Telescope 26003 AGA, Karl Storz SE \& Co. KG, Tuttlingen, Germany). This 5-megapixel camera has a resolution of 2448x2048 giving us optimal data for further \gls{3d} reconstructions.
As seen in Figure~\ref{fig:platform}, the laparoscope is mounted directly to the Kinova robotic arm using \gls{3d}-printed and laser-cut brackets. 
We consider two different light source settings for our study: the Storz laparoscopic light source (Storz D-LIGHT C 201336 20) attached directly to the laparoscope by fibre optic light cable (Storz 495 NCS) and the overhead ceiling-mounted surgical lights (Maquet Volista Surgical Light, Maquet GmBH, Rastatt, Germany). The overhead surgical light source evenly illuminates the image and eliminates vignetting effects visible when using the laparoscopic light source only. Whilst the surgical overhead light source is not realistic for \gls{mis}, it gives an ideal set of lighting conditions to compare to.

\subsection{Trajectories}
The aim of the robot's laparoscopic multi-view acquisition is to image the specimen organ from various angles, ensuring full surface coverage at least once. We represent this acquisition process as a sphere with the sample positioned at the centre. In Figure~\ref{fig:platform}.b, we depict this using two spherical scanning trajectories: Open-Close and Open-Far. The key difference between them is the variation in the distance to the specimen, where the Open-close distance ($d^{open-close}_{lap}$) is shorter than the Open-Far distance ($d^{open-far}_{lap}$). While this setup provides excellent coverage, it is only suitable for open surgery where there are no constraints on the surgical field. 

To achieve these spherical trajectories we set a \gls{rcm} colinear, but external to the  laparoscope and place it  at the centre of these virtual spheres. Thus the tip of the laparoscope moves in a spherical trajectory around the \gls{rcm} and all poses can be simply defined as an orientation of the \gls{rcm}.

In laparoscopic surgery, however, the endoscope's access is limited by the trocar — a single entry point through the skin surface. Therefore we define a third trajectory, the Trocar trajectory, to replicate this scenario. Although this is more clinically realistic, it introduces significant challenges for organ imaging, due to reduced \gls{fov}.


Our platform relies on MoveIt2 which utilises the \gls{ompl}~\cite{sucan2012open,kingston2019exploring} to plan these constrained trajectories, specifically using the RRTConnect planner. We use the default \gls{ik} solver in MoveIt2 which is the Orocos Kinematics and Dynamics plugin~\cite{kdl-url}. Given the non-uniqueness of \gls{rcm} pose solutions with a 7-\gls{dof}, we change the kinematics solver joint states seed to the current position rather than all zeroes to prevent awkward \gls{ik} solutions that require large movements from the current state.

\subsection{Pose Generation Using Fibonacci Sphere Sampling}
Achieving an even distribution of points on a spherical surface is essential for optimal coverage and accurate 3D reconstruction. Traditional methods, like uniformly sampling azimuthal and altitude angles, often result in uneven distributions, causing point clustering near the poles. This leads to sampling biases and inefficiencies, particularly in multi-view image acquisition for 3D reconstruction.
We employed the Fibonacci sphere sampling method to generate evenly distributed poses on a spherical surface~\cite{keinert2015spherical}. This approach provides a uniform distribution of points across a sphere, avoiding clustering near the poles as seen in Figure~\ref{fig:platform}.b.
The Fibonacci sphere is generated by calculating the azimuthal angle $\theta$ for each point on the sphere. This angle is incremented by the golden angle $\phi = \pi(3 - \sqrt{5})$, ensuring uniform spacing. The vertical position $y$ is distributed evenly between 1 and -1. 

To maintain focus on a region of interest and to allow adjustment for different OR settings, we limit the poses for all three trajectories to those within a specified angular range, defined by an angle limit $\theta_{max}$. This ensures that only points with an elevation angle greater than $90^\circ - \theta_{max}$ are included, thereby restricting poses to the top of the sphere. Finally, the orientation of each pose is computed by aligning the axis co-linear to the laparoscope with the vector pointing from the centre of the sphere to the sampled point.

\subsection{Data Recording}
We use our custom-developed Python app built on the FLIR Spinnaker \gls{sdk} to preview and capture data. The app can capture images and video frames. Each video frame is saved as a numpy file and frame information such as exposure time and current framerate is stored in a csv frame log. The \gls{rcm} and end-effector positions of the robot are also saved for analysis.

\subsection{Hand-Eye Calibration}
An additional dataset was captured on a ChAruco board using all three trajectories to ensure a good calibration. Each frame of this dataset underwent min-max normalisation to enhance the ChAruco board detection. The intrinsic calibration of the camera was obtained using OpenCV, with a reprojection error of 0.88 pixels. The robot was then hand-eye calibrated using a dual quaternion method~\cite{daniilidis1999hand} to obtain the transform from the end-effector of the robotic arm to the principal point of the camera.

\section{Data Acquisition}

For this work, we used a total of eight \textit{ex-vivo} ovine organs: six kidneys and two livers. \gls{rgb} frames from each organ set can be seen in Figure~\ref{fig:3Dpipeline}.a. Four of the kidneys were paired to give a total of six organ sets. Each organ was placed on a tray with a featured background to aid the \gls{sfm} pipeline.
Three different trajectories were used to image each organ set as visualised in Figure~\ref{fig:platform}:
\begin{enumerate}
    \item Trocar: the \gls{rcm} was placed $\sim$120~$mm$ above the centre of the sample. This leaves the tip $\sim$80~$mm$ above the sample creating a realistic MIS scenario
    \item Open-Close: the \gls{rcm} is placed in the sample with $d_{lap}$~=~80~$mm$  
    \item Open-Far: the \gls{rcm} is placed in the sample with \\$d_{lap}$~=~120~$mm$ for a wider \gls{fov}.
\end{enumerate}
The organs were imaged with two different light source configurations: laparoscopic light source only (Laparos.)  and a combination of Laparos. and the overhead surgical light source (Surgical) seen in Figure~\ref{fig:platform}. For the kidney datasets, each of the trajectories was performed with both light source configurations leading to six videos per organ set and twenty-four total videos. For the liver sample sets only the Trocar and Open-Far trajectory were performed, because the livers were larger the Open-Close trajectory was skipped to minimise the risk of collisions. This led to four videos per liver, eight in total and an overall dataset size of 32 videos.

In tackling issues with ground truth, as detailed in our related work, we implement a laser scanner to acquire highly accurate \gls{3d} meshes and point clouds from our samples. We use the Nikon H120 ModelMaker and MCAx S (Nikon Corporation, Tokyo, Japan) for this approach. The laser scanning system has an accuracy of down to 7$\mu$m~\footnote{\url{https://industry.nikon.com/en-us/products/3d-laser-scanners/manual-3d-scanning/modelmaker-h120/}} giving us the ability for to precisely evaluate our methods. This accuracy is greater than both structured light and the \gls{ct} ground truths supplied by current datasets as structured light/stereo depth estimation is limited by the resolution of the cameras and feature matching algorithms and \gls{ct} scanners have larger voxel sizes and requires a segmentation step for a \gls{3d} model for comparison that could introduce errors. A full \gls{3d} reconstruction ground truth does allow evaluation of a whole pipeline in comparison to solely depth maps, as projection from depth maps can introduce errors. Depth maps can be extracted from a ground truth point cloud provided camera poses and intrinsics, which does however rely on good registration from laser scanner to point cloud.



\section{3D Reconstruction}\label{sec:3drecon}
\begin{figure*}
    \centering
    \includegraphics[width=\linewidth]{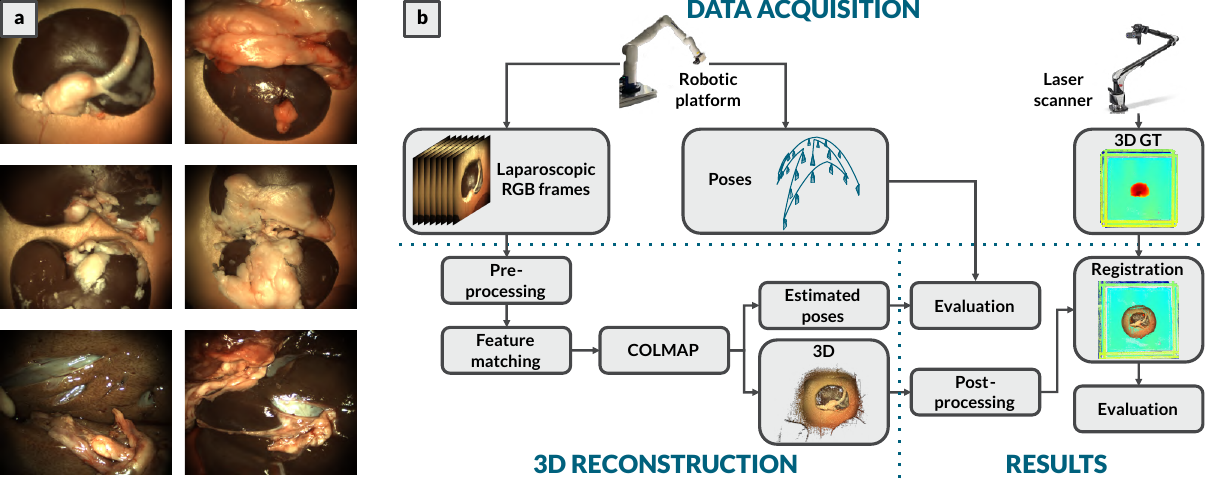}
    \caption{\textbf{a)} An \gls{rgb} frame from each of the 6 organ sample sets. Row 1\&2 show different kidney samples. In row 3 we show a section of two liver samples. \textbf{b)} Data acquisition: A summary of all the data collected by our platform. \gls{3d} Reconstruction: Our \gls{3d} reconstruction pipeline for processing our data. Results: Evaluation of the \gls{3d} reconstructions by comparing to acquired ground truth data.}
    \label{fig:3Dpipeline}
\end{figure*}

This section details our pipeline for producing \gls{3d} reconstructions from our robotic arm pipeline as seen in Figure~\ref{fig:3Dpipeline}. To perform the \gls{3d} reconstruction, we utilised two image feature extraction and matching methods, GIM-\gls{lg} and ALIKED-\gls{lg}. GIM-\gls{lg} uses a handcrafted feature extractor and a retrained matching component, \gls{lg}, following the GIM training framework. ALIKED is a lightweight CNN-based keypoint detector and descriptor extractor that we paired with the \gls{lg} matching component. These models were chosen due to their performance in~\cite{bonilla2024mismatched}, particularly in out-of-domain challenges. For comparison, we also paired \gls{lg} with SIFT~\cite{lowe2004sift}, a commonly used feature detector.
All the data was processed on a workstation with an AMD Ryzen Threadripper PRO 5975WX CPU and two NVIDIA RTX A6000 GPUs. Following this step we convert the data to standard COLMAP data structure and create sparse and dense reconstructions.

\subsubsection{Structure from Motion}
The images are undistorted using calibration data, and each video, with an average of 909 frames, is sub-sampled to 100 frames to reduce computational complexity due to high frame correlation. To generate a list of image pairs, we use the method from~\cite{bonilla2024mismatched} by pairing sequential frames using a sliding window. We enhance the matching process by utilising a pre-trained DINOv2-SALAD~\cite{izquierdo2024optimal} model to generate extra pairs. DINOv2-SALAD acts on each frame in the sub-sampled dataset and finds correspondences to other frames in the set. DINOv2 is employed for local feature extraction, while SALAD aggregates these local features to clusters and uses the optimal transport method to generate global features. Following this approach resulted in an average of 1169 pairs per dataset.
Subsequent image pairs were fed to the image matchers ALIKED-\gls{lg},GIM-\gls{lg} and SIFT-\gls{lg} which returned matches for each pair. We then utilise the incremental \gls{sfm} from COLMAP to create dense \gls{3d} reconstructions and estimated camera poses which can be evaluated in comparison to the ground truth. The hand-eye calibrated poses were not used in the pipeline as we use them for evaluation of the predicted poses.
 
\subsubsection{Post-processing}
Prior to comparison, the dense point clouds undergo a post-processing step. Due to our pipeline returning relative point clouds, the point clouds are manually registered to the laser scan ground truth using Open3D~\cite{Zhou2018} by manually selecting matching points in both point clouds before undergoing \gls{icp} registration with the laser scan to refine the registration. 
Manually registered point clouds are downsampled to 0.5~$mm$ and cleaned from statistical outliers using Open3D. The mean distance to its k nearest neighbours is calculated using a KD-Tree for efficient neighbour queries for every point. Subsequently, the global mean and standard deviation are computed for all points. Points are considered outliers if their mean neighbour displacements exceed a threshold defined as the population mean plus a multiple of the standard deviation. This method removes isolated noise while preserving the point cloud structure, making it suitable for denoising in \gls{3d} reconstruction workflows. For this, we set $\texttt{k = 20}$ and $\texttt{std\_ratio = 1.0}$ to maintain data quality. Finally, points more than 60~$mm$ from the centroid are removed.
Finally, we use Open3D's Point-to-Plane \gls{icp} registration~\cite{chen1992object} to register our reconstruction with the ground truth with the default Tukey Loss with $k=$ 1. This uses the normals of the target scan, obtained directly from the laser scanner, in its objective function to increase the fidelity of the \gls{icp} algorithm.

While the majority of the reconstructions were successful, 12 out of 96 were excluded due to issues in reconstruction with the SIFT-LG method account for 50\% of failures. Half of all exclusions were due to insufficient point cloud data resulting from too few features detected, and the remaining due to excessive noise, particularly in the trocar trajectory, which posed challenges for feature detection and surface reconstruction due to restricted views.

\section{RESULTS}

\begin{figure*}
    \centering
    \includegraphics[width=\linewidth]{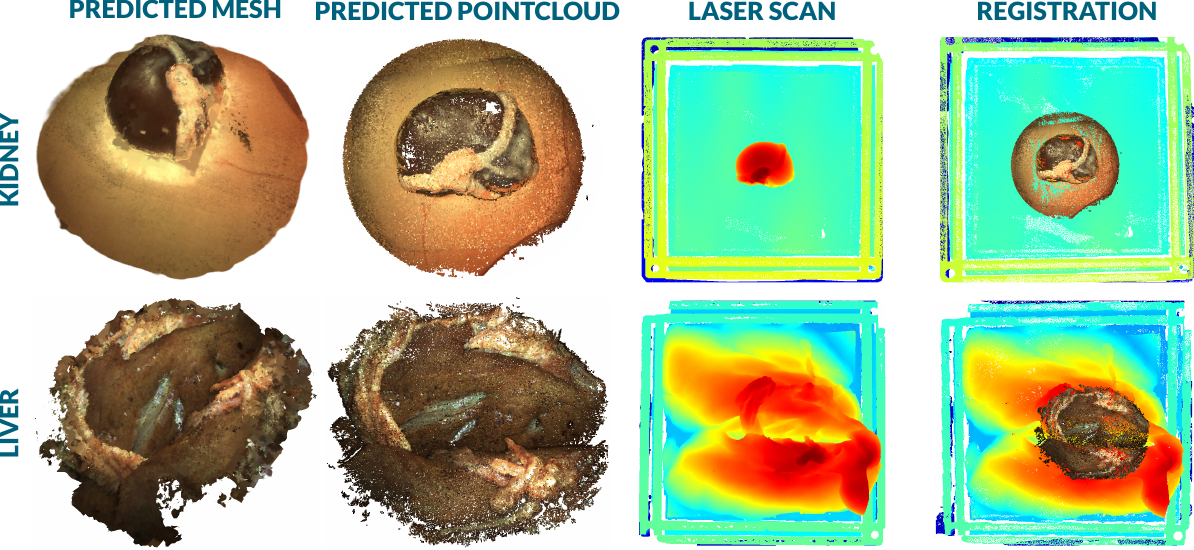}
    \caption{3D reconstructions of a kidney and liver obtained using different trajectories and lighting conditions. For the kidney (row 1), data was captured using the Open-Far trajectory under operating room lighting, and processed with the ALIKED-\gls{lg} image matcher. The liver (row 2) was captured using the trocar trajectory under laparoscopic lighting and processed using the GIM-\gls{lg} matcher. In both cases, the predicted \gls{3d} models (col 1\&2) are compared to the ground truth laser scans (col 3) and the post-processed, aligned reconstructions (col 4).}\label{fig:qualres}
\end{figure*}

\subsection{Qualitative Results}


In this section, we present our qualitative results and findings. In Figure~\ref{fig:qualres} two representative cases are shown for a liver and a kidney. The first case involves a kidney acquired using the Open-Far trajectory with overhead OR lighting, processed using the ALIKED-\gls{lg} matcher. The second case involves a liver acquired using the trocar trajectory with laparoscopic lighting, processed using the GIM-\gls{lg} matcher. Both methods resulted in visually accurate \gls{3d} reconstructions despite the differences in acquisition trajectories and lighting conditions.
While the overall reconstructions appear good, we noticed small holes in some areas of the point clouds. These gaps are minor and do not significantly affect how the \gls{3d} models can be used, but they may slightly reduce the visual quality. For improved visualisation, we used Poisson meshes, which helped fill in some of the gaps and gave us cleaner-looking models. By observing the location of these holes it appears that one of the main challenges in the \gls{3d} pipeline is dealing with the dark, smooth surfaces of the organ and specular highlights. These surfaces, especially under laparoscopic lighting, made it harder for the algorithms to detect enough matching features, particularly when we used the limited perspectives of the trocar trajectory. With fewer camera angles, there were more instances of occlusions and missing data compared to the wider Open-Close and Open-Far trajectories, where the broader \gls{fov} provided a clearer picture.
Despite these difficulties with unreconstructed surfaces, the registrations between our predicted models and the ground truth laser scans appear to be visually well aligned with only minor discrepancies.

\subsection{3D Reconstruction Results}

\subsubsection{Metrics}

In evaluating the performance of point cloud registration to a ground truth laser scan, we employed three widely used distance metrics in millimetres: Chamfer distance, Hausdorff distance, and \gls{rmse}.  Each metric provides a different aspect of the registration quality: Chamfer distance measures average minimum distances between points, Hausdorff distance captures maximum deviations, and \gls{rmse} reflects average point-wise errors. Together, they give us a comprehensive view of how closely our reconstructions match the ground truth.\\
Due to gaps and holes in the incomplete laser scan, we excluded the top 5\% of distances from the metric calculations. This step prevents artificially inflated errors from points in the source cloud that are not covered by the target cloud.\\
Tables~\ref{tab:quantitative_lighting} and~\ref{tab:quantitative_trajectories} presents the quantitative evaluation of our point cloud registrations against the ground truth laser scans. Table~\ref{tab:quantitative_lighting} compares the performance of different methods — SIFT-\gls{lg}, GIM-\gls{lg}, and ALIKED-\gls{lg} — across the two lighting conditions (operating room and laparoscopic lighting) for both kidney and liver datasets averaged across all trajectories.

\begin{table}
\centering
\caption{Average metrics using the Laparos. and Surgical light source. Best results highlighted in bold and best for each light source setting individually is underscored.}
\label{tab:quantitative_lighting}
\begin{tabular}{llcc} 
\hline
\textbf{Metric} & \textbf{Method} & \textbf{Surgical} & \textbf{Laparos.} \\ \hline
\multicolumn{4}{c}{\textbf{Kidneys}} \\ \hline
\multirow{3}{*}{Chamfer} & SIFT-LG & 0.760 & 0.983 \\
 & GIM-LG & \underline{\textbf{0.605}} & \underline{0.786} \\
 & \multicolumn{1}{c}{ALIKED-LG} & 0.872 & 0.869 \\ \hline
\multirow{3}{*}{Hausdorff} & SIFT-LG & 2.626 & 3.783 \\
 & GIM-LG & \textbf{\underline{2.017}} & \underline{2.922} \\
 & \multicolumn{1}{c}{ALIKED-LG} & 3.090 & 3.261 \\ \hline
\multirow{3}{*}{RMSE} & SIFT-LG & 0.995 & 1.294 \\ 
 & GIM-LG & \underline{\textbf{0.764}} & \underline{1.029} \\
 & \multicolumn{1}{c}{ALIKED-LG} & 1.161 & 1.146 \\ \hline
\multicolumn{4}{c}{\textbf{Livers }} \\ \hline
\multirow{3}{*}{Chamfer} & SIFT-LG & 1.571 & 1.735 \\
 & GIM-LG & 0.686 & 1.225 \\
 & ALIKED-LG & \underline{\textbf{0.551}} & \underline{0.588} \\ \hline
\multirow{3}{*}{Hausdorff} & SIFT-LG & 5.430 & 6.072 \\
 & GIM-LG & 2.499 & 4.347 \\
 & ALIKED-LG & \underline{\textbf{2.030}} & \underline{2.244}\\ \hline
\multirow{3}{*}{RMSE} & SIFT-LG & 2.068 & 2.284 \\
 & GIM-LG & 0.892 & 1.588 \\
 & ALIKED-LG & \underline{\textbf{0.720}} & \underline{0.769} \\
\hline
\end{tabular}
\end{table}

\subsubsection{Differences in Organ Reconstructions}
In table~\ref{tab:quantitative_lighting} we show the results of the organ 3D reconstruction for the different methods and datasets. Overall the reconstruction results for the different organs and methods show consistent finding with ALIKED-LG and GIM-LG outperforming the SIFT-LG baseline method consistently. Only for the Kidney dataset under the surgical light setting the SIFT-LG shows better Chamfer results (-0.112) and \gls{rmse} (-0.166) compared to the ALIKED-LG method. The sharper transitions and irregularities at the kidney edges make accurate feature matching often more difficult and ALIKED-LG seems to struggle in these scenarios. On the Liver dataset we see that the ALIKED-LG model outperforms GIM-LG and SIFT-LG by a large margin. 

\subsubsection{Lighting Conditions}
Table~\ref{tab:quantitative_lighting} highlights the influence of lighting conditions on registration metrics. Surgical lighting, offering broader and more even illumination, generally produced better results, particularly with lower Chamfer, Hausdorff, and \gls{rmse} values. In contrast, laparoscopic lighting, which is more focused and directional, led to slightly higher errors, especially in the Hausdorff distance, likely due to the introduction of shadows, uneven illumination, and pronounced vignetting. Despite these differences, both GIM-\gls{lg} and ALIKED-\gls{lg} exhibited strong performance across lighting conditions, with GIM-\gls{lg} performing slightly better overall, particularly in the Chamfer (2.017 and 2.922) and \gls{rmse} (0.764 and 1.029) metrics on the kidney dataset. These results confirm that both methods are effective for \gls{3d} reconstruction in surgical environments, with GIM-\gls{lg} showing especially consistent performance, without outliers, even in challenging lighting conditions.

\subsubsection{Trajectories}

\begin{table}
\setlength{\tabcolsep}{2pt}
\centering
\caption{Average metrics across trajectories for kidneys and livers. The best-performing metrics for each organ across all Methods and Trajectories are highlighted in bold.}
\label{tab:quantitative_trajectories}
\begin{tabular}{llccc} 
\hline
\textbf{Metric} & \textbf{Method} & \textbf{Trocar} & \textbf{Open-close} & \textbf{Open-far} \\ 
\hline
\multicolumn{5}{c}{\textbf{Kidneys}} \\ 
\hline
\multirow{3}{*}{Chamfer} & SIFT-\gls{lg}   & 0.974 & 0.787 & 0.855 \\
                         & GIM-\gls{lg}    & 0.844 & 0.669 & \textbf{0.581} \\
                         & ALIKED-\gls{lg} & 0.815 & 0.801& 0.981 \\
\hline
\multirow{3}{*}{Hausdorff} & SIFT-\gls{lg}   & 3.363 & 3.158 & 3.093 \\
                           & GIM-\gls{lg}    & 2.814 & 2.651 & \textbf{1.930} \\
                           & ALIKED-\gls{lg} & 2.972 & 3.051 & 3.452 \\
\hline
\multirow{3}{*}{RMSE} & SIFT-\gls{lg}   & 1.246 & 1.041 & 1.147 \\
                      & GIM-\gls{lg}    & 1.078 & 0.882 & \textbf{0.735} \\
                      & ALIKED-\gls{lg} & 1.059 & 1.058 & 1.320 \\
\hline
\multicolumn{5}{c}{\textbf{Livers}} \\ 
\hline
\multirow{3}{*}{Chamfer} & SIFT-\gls{lg}   & 1.990 & -     & 0.550 \\
                         & GIM-\gls{lg}    & 0.757 & -     & 1.172 \\
                         & ALIKED-\gls{lg} & 0.692 & -     & \textbf{0.483} \\
\hline
\multirow{3}{*}{Hausdorff} & SIFT-\gls{lg}   & 6.902 & -     & 2.112 \\
                           & GIM-\gls{lg}    & 2.721 & -     & 4.181 \\
                           & ALIKED-\gls{lg} & 2.391 & -     & \textbf{1.973} \\
\hline
\multirow{3}{*}{RMSE} & SIFT-\gls{lg}   & 2.621 & -     & 0.720 \\
                      & GIM-\gls{lg}    & 0.969 & -     & 1.530 \\
                      & ALIKED-\gls{lg} & 0.876 & -     & \textbf{0.653} \\
\hline
\end{tabular}
\end{table}

Table~\ref{tab:quantitative_trajectories} presents the average metrics for each trajectory. Due to the larger size of the livers, no open-close trajectory was recorded to avoid collisions with the organ. The metric differences between the Open-close and Open-far datasets highlight that the choice of $d_{lap}$ significantly impacts reconstruction quality, emphasizing the need for careful parameter selection. Among the trajectories, the trocar trajectory—being the most realistic—performed the worst, likely due to the limited number of views. However, when SIFT data was excluded, the average Chamfer distances across all organs were 0.797, 0.735, and 0.795 for the trocar, open-close, and open-far trajectories, respectively. This demonstrates that modern feature matching techniques can achieve comparable results even with realistic, patient-specific trajectories.

\subsection{Pose Evaluation}

\begin{figure}
    \centering
    \includegraphics[width=\linewidth]{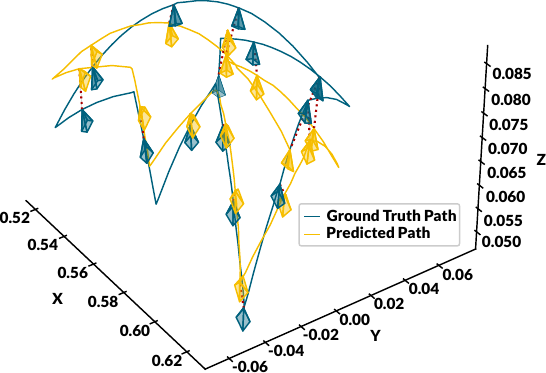}
    \caption{A graph depicting a sub-sampled set of poses of the ground truth (blue) and predicted (yellow) poses for a kidney dataset with the OR light source and the open-far trajectory using the ALIKED-LG image matcher.}
    \label{fig:poses}
\end{figure}

We employed the Umeyama algorithm~\cite{umeyama1991least} to align our predicted poses with the ground truth for comparison. Figure~\ref{fig:qualres} illustrates a subsampled set of pose estimations from a kidney dataset using an open-close trajectory, with the ALIKED-\gls{lg} feature matching version of the pipeline. Overall, the poses were accurate, with translation estimations appearing slightly better than rotations. The \gls{rpe} for rotation was 0.0161 radians using ALIKED-\gls{lg}, 0.0032 for GIM-\gls{lg}, and 0.0290. For translations, the \gls{rpe} was 0.265~$mm$ for ALIKED-\gls{lg}, 0.285~$mm$, and 0.245~$mm$ for GIM-\gls{lg}.

On average, 88.78\% of frames were predicted, with a standard deviation of 27.85\%, which can be attributed to a few anomalous cases with fewer than 15 frames having predicted poses. Across all datasets, 81.25\% successfully predicted poses for all frames, with GIM-\gls{lg} slightly outperforming ALIKED-\gls{lg} at 90.95\% compared to 86.60\%. This highlights the robustness of both methods, especially GIM-\gls{lg}, in feature matching and pose prediction.

\section{CONCLUSIONS}

In this work, we have customized a robotic arm platform for medical applications, demonstrated a versatile acquisition protocol, and utilized multi-view images for 3D reconstructions. Our platform has proven effective for efficiently and reliably collecting extensive datasets, streamlining the data acquisition process while ensuring high realism and adaptability through relevant metrics. This platform holds significant potential for advancing research, particularly in challenging environments involving smoke, blood, tool occlusions, and deformable tissue. Future developments may focus on adapting the platform to these complex surgical scenarios, with an emphasis on real-time processing and further enhancement of data quality.

\bibliography{root,bibtex}
\bibliographystyle{IEEEtran}

\end{document}